\documentclass[krantz1,ChapterTOCs]{krantz} 
\usepackage{fixltx2e,fix-cm}
\usepackage{amssymb}
\usepackage{amsmath}
\usepackage{graphicx}
\usepackage{subfigure}
\usepackage{makeidx}
\usepackage{multicol}
\usepackage{multirow}
\usepackage{array}
\usepackage{adjustbox}

\makeindex

{\par\addvspace{6pt}\normalfont {\bfseries #1}\hskip\labelsep\ignorespaces\itshape}
{\par\addvspace{6pt}}

\begin{document}



\cleardoublepage
\setcounter{page}{7} 
\tableofcontents

\mainmatter

\chapterauthor{Kaidong Li, Wenchi Ma, Usman Sajid, Yuanwei Wu, Guanghui Wang}{Department of Electrical Engineering and Computer Science, University of Kansas, Lawrence, KS 66045}

\chapter{Object Detection with Convolutional Neural Networks}



\section{Introduction}\label{intro}

Deep learning was first proposed in 2006 \cite{hinton2006fast}, however, it did not attract much attention until 2012. With the development of computational power and a large amount of labeled datasets \cite{bengio2013representation, arel2010deep}, deep learning has been proven to be very effective in extracting intrinsic structure and high-level features. Significant progress has been made in solving various problems in the artificial intelligence community \cite{lecun2015deep}, especially in areas where the data are multi-dimensional and the features are difficult to hand-engineer, including speech recognition \cite{hinton2012deep, sainath2013deep, dahl2012context}, natural language processing \cite{collobert2011natural, sutskever2014sequence, socher2011parsing} and computer vision \cite{krizhevsky2012imagenet, szegedy2015going, he2018learning, gao2016novel}. Deep learning also shows dominance on areas like business analysis \cite{najafabadi2015deep}, medical diagnostic \cite{litjens2017survey, mo2018efficient}, art creation \cite{elgammal2017can} and image translation \cite{Xu2019}\cite{Xu2019b}. Since little human input is required, deep learning will continue to make more and more impact in the future with the increase of computing power and exploding data growth.

Among all these areas, computer vision has witnessed remarkable successful applications of deep learning \cite{ma2018mdcn, zhang2018bpgrad, liu2018deep}. Vision, as the most important sense in terms of navigation and recognition for humans, provides the most information about the surroundings. Thanks to the rapid development of digital cameras in the last 10 years, image sensors have become one of the most accessible hardware. Researchers tend to explore the possibility of using computer vision in applications like autonomous driving, visual surveillance, facial recognition \cite{cen2019dictionary}, etc. The demands in these areas, in return, draw tremendous attention and provide huge resources to the computer vision field. As a result, it is developing at a speed that has been rarely seen in other research fields. Every month, sometimes even within a week, we have seen record performance achieved on datasets like ImageNet \cite{krizhevsky2012imagenet}, PASCAL VOC \cite{everingham2010pascal} and COCO \cite{lin2014microsoft}.

Object detection is a fundamental step for many computer vision applications \cite{liu2018deep,bharati2016fast,wu2017vision,bharati2018real}. The architecture of convolutional neural networks (CNN) designed for object detection are usually used as the first step to extract objects' spatial and classification information. Then more modules are added for specific applications. For example, tracking shares a very similar task to object detection with the addition of a temporal module. Therefore, the performance of object detection will affect almost all other computer vision research. A huge amount of effort has been put into its improvements \cite{wei2018quantization, kang2018t, chu2018deep}.

In this chapter, we present a brief overview of the recent development in object detection using convolutional neural networks (CNN). Several classical CNN-based detectors are presented. Some developments are based on the detector architectures \cite{ren2015faster, redmon2016you}, while others are focused on solving certain problems, like model degradation and small-scale object detection \cite{he2016deep, lin2018focal, zhao2019M2DetAS}. The chapter also presents some performance comparison results of different models on several benchmark datasets. Through the discussion of these models, we hope to give readers a general idea about the developments of CNN-based object detection. 

\subsection{Major Evolution}
To talk about object detection, we have to mention image classification. Image classification is the task of assigning a label to an input image from a fixed set of categories. The assigned label is usually corresponding to the most salient object in the image. It works best when the object is centered and dominating in the image frame. However, most images contain multiple objects, scattering in the frame with different scales. One label is far from enough to describe the contextual meaning of images. Therefore, object detection is introduced to not only output multiple labels corresponding to an image but also generate the spatial region associated with each label.

\textbf{Two-stage models.} Inspired by image classification, it is natural for researchers to detect the objects by exploring a two-stage approach. In most two-stage models, the first stage is for region proposal, followed by image classification on the proposed regions. With this modification, additional work is done on localization, with little changes on previous classification models. For the region proposal part, some systems employ a sliding window technique, like Deformable Parts Models (DPM) \cite{felzenszwalb2008discriminatively} and OverFeat \cite{sermanet2013overfeat}. With certain strategies, these methods usually apply classifiers on windows at different locations with different scales. Another region proposal is Selective Search \cite{uijlings2013selective}, which is adopted by R-CNN \cite{girshick2014rich}. R-CNN selectively extracts around 2,000 bottom-up region proposals \cite{girshick2014rich}, which greatly reduces the regions needed in sliding window methods.

The two-stage models, generally speaking, yield higher accuracy. With each stage doing its specific task, these models perform better on objects with various sizes. However, with the demands for real-time detection, two-stage models show its weakness in processing speed.

\textbf{One-stage models.} With more and more powerful computing ability available, CNN layers become deeper and deeper. Researchers are able to utilize one-stage methods, like YOLO (You Only Look Once) \cite{redmon2016you} and SSD (Single Shot Multibox Detector) \cite{liu2016ssd}, with faster detection speed and similar or sometimes even higher accuracy. One-stage models usually divide the image into \(N \times N\) grids. Each grid cell is responsible for the object whose center falls into that grid. Thus, the output is an \(N \times N \times S\) tensor. Each of the \(S\) feature maps is an \(N \times N\) matrix, with each element describes the feature on its corresponding grid cell. A common design of these feature maps for one object consists of the following \(S = (5 + C)\) values:

\begin{itemize} \setlength\itemsep{-3pt}

\item Four values for the bounding box dimensions (x coordinate, y coordinate, height, and width);
\item One value for the possibility of this grid cell containing the object;
\item \(C\) values indicating which class this object belongs to.
\end{itemize}

One-stage models, thanks to its simpler architecture, usually require less computational resource. Some recent networks could achieve more than 150 frames per second (fps) \cite{redmon2018yolov3}. However, the tread-off is accuracy, especially for small scale objects as shown in Table~\ref{table:vocTable} in Chapter~\ref{subsec:vocdata}. Another drawback is the class imbalance during training. In order to detect \(B\) objects in a grid, the output tensor we mentioned above include information for \(N \times N \times B\) number of anchor boxes. But among them, only a few contain objects. The ratio most of the time is about 1000:1 \cite{lin2018focal}. This results in low efficiency during training.

\subsection{Other Development}
With the general architecture established, new detector development mainly focuses on certain aspects to improve the performance. 

Studies \cite{simonyan2014very, szegedy2015going} show strong evidence that network depth is crucial in CNN model performance. However, when we expect the model to converge for a loss function during training, the problem of vanishing/exploding gradients prevents it from behaving this way. The study of Bradley (2009) found that the back-propagated gradients were smaller as one moves from the output layer towards the input layer, just after initialization \cite{glorot2010understanding}. Therefore, training becomes inefficient towards the input layers. This shows the difficulty of training very deep models. On the other hand, research shows that simply adding additional layers will not necessarily result in better detection performance. Monitoring the training process shows added layers are trained to become identity maps. Therefore, it can only generate models whose performance is equal to a shallower network at most after certain amount of layers. To address this issue, skip connections are introduced in the networks \cite{he2016deep, huang2017densely, chen2017dual} to pass information between nonadjacent layers. 

Feature pyramid networks (FPN) \cite{lin2017feature}, which outputs feature maps at different scales, can detect objects at very different scales. This idea could also be found when the features were hand-engineered \cite{lowe1999object}. Another problem is the trailing accuracy in one-stage detector. According to Lin {\it et al.} \cite{lin2018focal}, the reason for lower accuracy is caused by extreme foreground-background class imbalance during training. To address this issue, Lin {\it et al.} introduced RetinaNet \cite{lin2018focal} by defining focal loss to reduce the weight of background loss.

\section{Two-Stage Model}\label{two stage models}
A two-stage model usually consists of regional proposals extraction and classification. It is an intuitive idea to do after the success of image classification. This type of model could use the proven image classification network after region proposal stage. In addition, the two steps also, to some extent, resemble how humans receive visual information and arrange attentions on regions of interest (ROI). In this section, we will introduce the R-CNN, fast R-CNN, and faster R-CNN models \cite{girshick2014rich, girshick2015fast, ren2015faster}, and  discuss the improvements of each model.

\subsection{Regions with CNN Features (R-CNN)}
\begin{figure}
	\includegraphics[height=122pt]{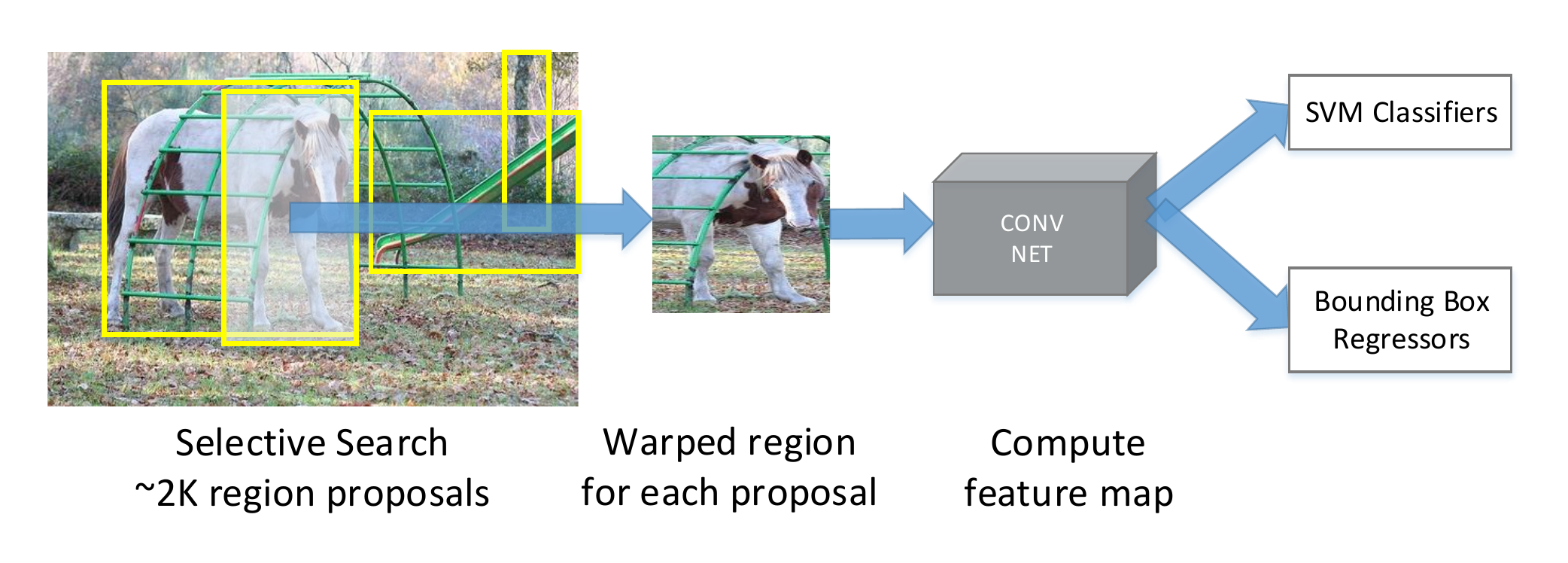}
	\centering
	\caption{\textbf{An Overview of the R-CNN detection system.} The system generates 2k proposals. Each warped proposal produces feature maps. Then the system classifies each proposal and refines the bounding box prediction.}
	\label{rcnnOverview}
\end{figure}

R-CNN was developed using a multi-stage process, which is shown in Figure~\ref{rcnnOverview}. It generally can be divided into two stages. 

The first stage is the region proposal extraction. The R-CNN model utilizes selective search \cite{uijlings2013selective}, which takes the entire image as input and generates around 2,000 class-independent region proposals. In theory, R-CNN is able to work with any region proposal methods. Selective search is chosen because it performs well and has been employed by other detection models. 

The second stage starts with a CNN that takes a fixed-dimension image as input and generates a fixed-length feature vector as output. The input comes from the regional proposals. Each proposal is wrapped into the required size regardless of its original size and aspect ratio. Then using the CNN feature vector, pre-trained class-specific linear SVMs are applied to calculate the class scores. Girshick {\it et al.} \cite{girshick2014rich} conducted an error analysis, and based on the analysis result, a bounding-box regression is added to reduce the localization errors using the CNN feature vector. The regressor, which is class specific, can refine bounding box prediction.

\subsection{Fast R-CNN}

R-CNN \cite{girshick2015fast} has achieved great improvement compared to the previous best methods. Let us take the results on PASCAL VOC 2012 as an example. R-CNN has a 30\% relative performance increase over the previous best algorithm. Apart from all the progress, R-CNN suffers from two major limitations. The first one is the space and time cost in training. In the second stage, the features for the \~{}2,000 region proposals are all extracted separately using CNN and stored to the disk. With VGG16, VOC07 trainval dataset of 5,000 images requires 2.5 GPU-days \footnote{a unit of computational complexity equivalent to that a single GPU can complete in a day}, and the memory expands to hundreds of GB space. The second limitation is the slow detection speed. For each test image, it takes a GPU 47 seconds with VGG16. This weakness prevents the R-CNN model from real-time applications.

\begin{figure}
	\includegraphics[height=122pt]{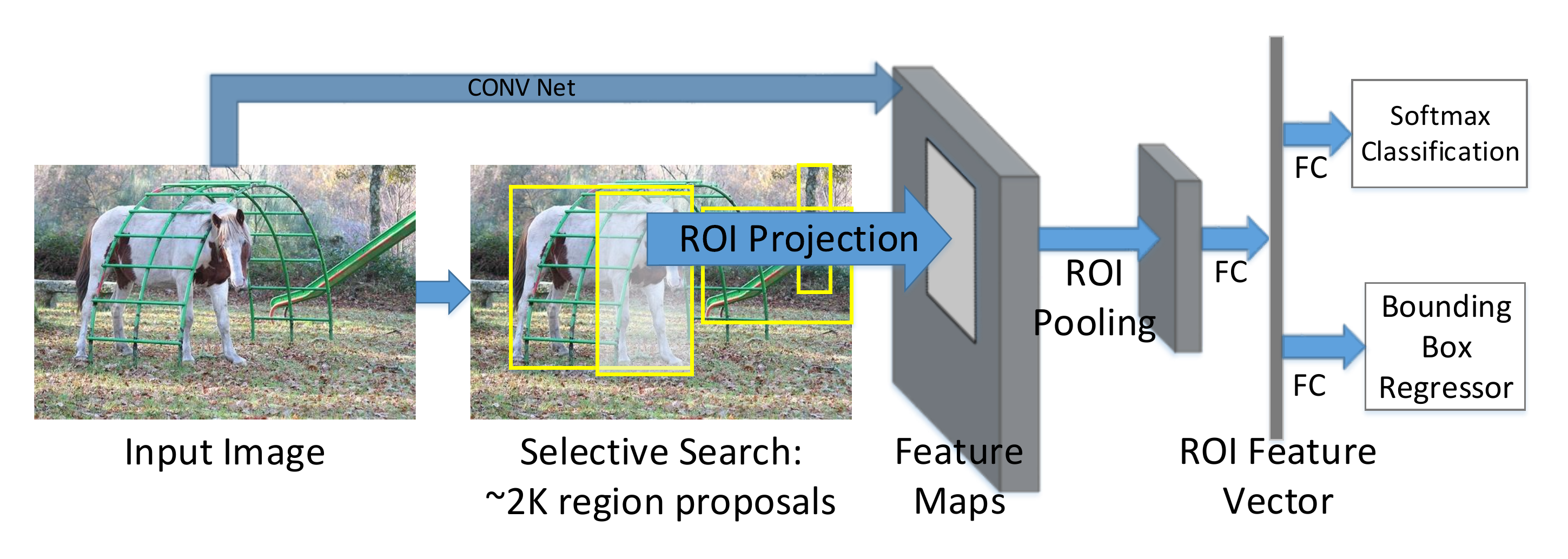}
	\centering
	\caption{\textbf{An overview of Fast R-CNN structure.} Feature maps of the entire image are extracted. ROI is projected to the feature maps and pooled into a fixed size. Then ROI feature vector is computed using fully connected layers. Using the vector, softmax classification probability and bounding box regression are calculated as outputs for each proposal.}
	\label{fastRcnnOverview}
\end{figure}

From the analysis above, both of the problems are caused by individually calculating a huge number of feature vectors for region proposals. Fast R-CNN is designed to overcome this inefficiency. The architecture of fast R-CNN is illustrated in Figure~\ref{fastRcnnOverview}. The first stage remains unchanged, and generates region proposals using selective search. For the second stage, instead of generating feature vectors for each region proposals separately, fast R-CNN first processes the entire image using CNN to calculate one feature map. Then, for each object proposal, a region of interest (RoI) pooling layer extracts a fixed-length feature vector from the feature map \cite{girshick2015fast}, and each RoI feature vector is processed by a sequence of fully connected layers and forked into two branches. The first branch calculates class scores with a softmax layer; and the other branch is again a bounding-box regressor which refines the bounding-box estimation.

In fast R-CNN, the RoI pooling layer is the key part where the region proposals are translated into fixed-dimension feature vectors. This layer takes the feature map of the entire image and RoI as input. Each RoI is defined by a four-tuple \((r, c, h, w)\) that specifies its top-left corner \((r, c)\) and its height and width \((h, w)\) \cite{girshick2015fast}. Then, the \((h, w)\) dimension of RoI is divided into an \(H \times W\) grid, where \(H\) and \(W\) are layer hyper-parameter. Standard max pooling is conducted to each grid cell in each feature map channel.

With similar mAP (mean average precision), fast R-CNN is very successful in improving the training/testing speed. For large scale objects defined in \cite{he2016deep}, fast R-CNN is 8.8 times faster in training and 146 times faster in detection \cite{girshick2015fast}. It achieves 0.32s/image performance, which is a significant leap towards real-time detection. 

\subsection{Faster R-CNN}

To further increase the detection speed, researchers discover that the region proposal computation is the bottleneck of performance improvement. Before fast R-CNN, the time taken at the second stage is significantly more than that at the region proposal stage. Considering selective search \cite{uijlings2013selective} works robustly and is a very popular method, improving region proposal seems unnecessary. However, with the improvement in fast R-CNN, selective search now is an order slower compared to classification stage. Even with some proposal methods that are balanced between speed and quality, region proposal stage still costs as much as the second stage. To solve this issue, Ren {\it et al.} \cite{ren2015faster} introduced a Region Proposal Network (RPN), which shares the feature map of the entire image with stage two. 

\begin{figure}
	\includegraphics[height=250pt]{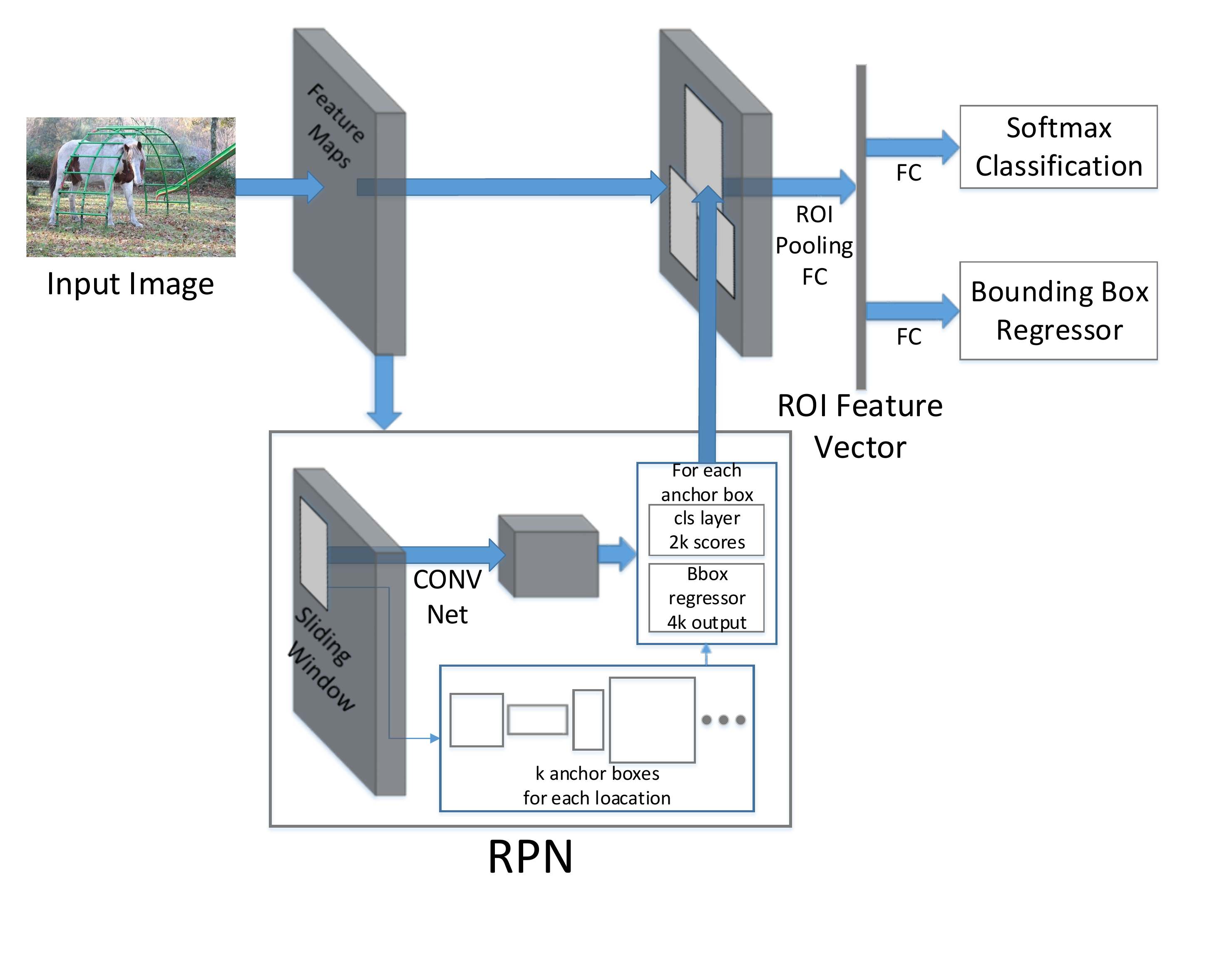}
	\centering
	\caption{\textbf{An overview of the Faster R-CNN framework.} Region proposal network (RPN) shares same feature maps with a fast R-CNN network; a small convolutional network slides over the shared feature maps to produce region proposal; and each location has k anchor boxes.}
	\label{fasterRcnnOverview}
\end{figure}

The first step in Faster R-CNN is a shared CNN network. The RPN then takes the shared feature maps as input and generates region proposals with scores indicating how confident the network is that there are objects in them. In the RPN, a small network will slide over the input feature map as indicated in Figure~\ref{fasterRcnnOverview}. Each sliding window will generate a lower-dimensional vector. This lower-dimensional vector will be the input of two sibling fully-connected layers, a bounding-box regression layer (reg layer) and a bounding-box classification layer (cls layer). To increase the non-linearity, ReLUs are applied to the output of the CNN layer. For each sliding window, $k$ region proposals will be generated. Correspondingly, the reg layer will have $4k$ outputs for bounding-box coordinates, and the cls layer will have $2k$ probabilities of object/not-object for each bounding-box proposals. The $k$ proposals are parameterized relative to $k$ reference boxes, called anchors. Each anchor is centered at the sliding window in question \cite{ren2015faster}. The parameters includes the scales and aspect ratio. 

To train RPN, binary class labels are assigned to each anchor. Positive is assigned to two situations.

\begin{itemize} \setlength\itemsep{-3pt}
	\item Anchors with highest Intersection-over-Union (IoU) with ground-truth boxes;
	\item Anchors with IoU over 0.7 with ground-truth boxes.
\end{itemize}

Non-positive is given to anchors with IoU lower than 0.3 for all ground-truth boxes. During training, only positive or non-positive anchors contribute to loss functions. With the introduction of RPN, the total time to complete object detection on GPU is 198 ms using VGG as the CNN layer. Compared to the selective search, it is almost 10 times faster. The proposal stage is improved from 1,510 ms to only 10 ms. Combined, the new faster R-CNN achieves 5 fps.

Two-stage detectors usually have higher accuracy in comparison to one-stage detectors. Most of the models at the top detection dataset leader board are two-stage models. Region proposals with regressor to refine localization means they inherently produce much better results for bounding-box predictions. In terms of speed, these detectors were far from real-time performance when they were first introduced. With recent development, they are getting closer by simplifying the architecture. It evolves from models with multiple computationally heavy steps to models with a single shared CNN. After the faster R-CNN \cite{ren2015faster} was published, the networks are mainly based on shared feature maps. For example, R-FCN (region-based fully convolutional networks) \cite{dai2016r} and Cascade R-CNN \cite{cai2018cascade} focus on localization accuracy, and at the same time, they can both achieve around 10 fps. Region based two-stage detectors are becoming not only more accurate but also faster. Therefore, it becomes one of the main branch for object detection.

\section{One-Stage Model}\label{one stage model}
From the development of R-CNN, we can observe the success of sharing CNN in almost all the steps of a detection network. Region proposal, classification, and bounding-box refinement all take the output of the last shared CNN. This indicates that, if designed properly, the feature maps contains information about all above tasks. It is natural to explore the possibility to implement detection as a single regression problem.

\subsection{You Only Look Once (YOLO)}

YOLO \cite{redmon2016you} is a simple model that takes the entire image as input and simultaneously generates multiple bounding-boxes and class probabilities. Since YOLO takes input from the entire image, it reasons globally for detection, which means each detection takes the background and other objects into consideration. 

YOLO divides input images into \(S \times S\) grid. If an object center falls into a grid cell, this grid cell is responsible for that object’s detection. Each grid cell will predict \(B\) bounding boxes, and the bounding boxes are described by five parameters: \(x, y, w, h\), and the \(confidence\), where \((x, y)\) represents the center of the box relative to the bounds of the grid cell, $(w, h)$ are width and height relative to the dimensions of the whole image, respectively. The confidence is defined as:

$$Pr(Object) * IOU_{pred}^{truth}$$
which reflects how confident the prediction is that an object exists in this bounding box and how accurate the bounding box is relative to the ground truth box. The final parameter is \(C\) conditional class probabilities. It is the conditional probabilities of the grid cell containing an object, regardless of how many bounding boxes a grid cell has. Therefore, the final predictions are an \(S \times S \times (B \times 5 + C)\) tensor.

The network design is shown in Table~\ref{table:yoloArch}. It is inspired by GoogLeNet \cite{szegedy2015going}, followed by 2 fully connected layers: FC1 of 4,096 dimensional vector, and the final output FC2 of 7x7x30 tensor. In the paper, the authors also introduced a fast version with only 79 convolutional layers.

\begin{table}
\small
\centering
\caption{The architecture of the convolutional layers of YOLO \cite{redmon2016you}.}
\begin{adjustbox}{width=\textwidth}
\begin{tabular}{ r | r  r  r  r  r  l  r  l  r }
	stage  & image & conv1 & conv2 & conv3 & conv4 & & conv5 & & conv6 \\ 
	\hline
	output &\(448 \times 448\)&\(112 \times 112\)&\(56 \times 56\)&\(28 \times 28\)&\(14 \times 14\)& &\(7 \times 7\)& &\(7 \times 7\) \\ 
	~& 3 & 192 &256 &512 &1024 & & 1024 & & 1024 \\ 
	\hline
	size    & &\(7 \times 7\)&\(3 \times 3\)&\(1 \times 1\)&\(1 \times 1\)& \multirow{4}{*}{ \Bigg\} $\times$ 4} &\(1 \times 1\)& \multirow{4}{*}{ \Bigg\} $\times$ 2} & \(3 \times 3\) \\ 
	filter/ & &          64/2&           192&           128&          256 & &           512& &    1024        \\ 
	stride  & &              &              &\(3 \times 3\)&\(3 \times 3\)& &\(3 \times 3\)& & \(3 \times 3\) \\ 
	&         &              &              &           256&          512 & &          1024& &  1024          \\ 
	&         &              &              &\(1 \times 1\)&\(1 \times 1\)& &\(3 \times 3\)& &                \\ 
	&         &              &              &           256&          512 & &          1024& &                \\  
	&         &              &              &\(3 \times 3\)&\(3 \times 3\)& &\(3 \times 3\)& &                \\ 
	&         &              &              &           512&          1024& &        1024/2& &                \\ 
	\hline
	Maxpool & &\(2 \times 2\)&\(2 \times 2\)&\(2 \times 2\)&\(2 \times 2\)& &              & &                \\ 
	stride  & &             2&             2&             2&             2& &              & &                \\ 
	 
\end{tabular}
\end{adjustbox}
\label{table:yoloArch}
\end{table}

During training, the loss function is shown in Equation~\ref{yoloLoss}, where \(1_i^{obj}\) denotes if an object appears in cell \(i\), and \(1_{ij}^{obj}\) denotes that the $j$th bounding box predictor in cell $i$ is “responsible” for that prediction \cite{redmon2016you}. From the function, we can see that it only penalizes the error when an object exists and when a prediction is actually responsible for the ground truth.
\begin{equation}
\label{yoloLoss}
Loss = {Loss}_{bounding\ box} + {Loss}_{confidence} + {Loss}_{classification} 
\end{equation}
where
\begin{equation*}
\begin{split}
{Loss}_{bounding\ box} = & {\lambda}_{coord}{\sum}_{i=0}^{S^2}{\sum}_{j=0}^{B} 1_{ij}^{obj} [(x_i-\hat{x}_i)^2+(y_i-\hat{y}_i)^2 ] \\
&+{\lambda}_{coord}{\sum}_{i=0}^{S^2}{\sum}_{j=0}^{B} 1_{ij}^{obj}[(\sqrt{w_i}-\sqrt{\hat{w}_i} )^2 + (\sqrt{y_i}-\sqrt{\hat{y}_i} )^2 ],
\end{split}
\end{equation*}

\begin{equation*}
\begin{split}
{Loss}_{confidence} = &{\sum}_{i=0}^{S^2}{\sum}_{j=0}^{B}1_{ij}^{obj}(C_i-\hat{C}_i)^2 \\
&+{\lambda}_{noodj}{\sum}_{i=0}^{S^2}{\sum}_{j=0}^{B}1_{ij}^{obj}(C_i-\hat{C}_i )^2,
\end{split}
\end{equation*}

\begin{equation*}
\begin{split}
{Loss}_{classification} = &{\sum}_{i=0}^{S^2}1_{i}^{obj}{\sum}_{c \in classes}(p_i(c)-\hat{p}_i(c))^2.
\end{split}
\end{equation*}

\subsection{YOLOv2 and YOLO9000}
YOLOv2 \cite{redmon2017yolo9000} is an improved model based on YOLO. The original YOLO backbone network is replaced by a simpler Darknet-19, and the fully connected layers at the end are removed. Redmon {\it et al.} also tested other different design changes and only applied modifications with accuracy increases. YOLO9000 \cite{redmon2017yolo9000}, as the name suggests, can detect 9,000 object categories. Based on a slightly modified version of YOLOv2, this is achieved with a jointly training strategy on classification and detection datasets. A WorldTree hierarchy \cite{redmon2016you} is used to merge the ground truth classes from different datasets.

In this section, we will discuss some of the most effective modifications. Batch normalization can help to reduce the effect of internal covariate shifts \cite{ioffe2015batch}, thus accelerates convergence during training. By adding batch normalization to all CNN layers, the accuracy is improved by 2.4\%.

In YOLO, the classifer is trained on the resolution of $224 \times 224$. At the stage of detection, the resolution is increased to $448 \times 448$. In YOLOv2, during the last 10 epochs of classifier training, the image is changed to a full 448 resolution. So the detection training could focus on object detection rather than adapting to the new resolution. This gives a 4\% mAP increase. 
While trying anchor boxes with YOLO, the issue of instability is exposed. YOLO predicts the  box by generating offsets to each anchor. The most efficient training is when all objects are predicted by the closest anchor with a minimal amount of offsets. However, without offsets constrains, an original anchor could predict an object at any location. Therefore in YOLOv2 \cite{redmon2017yolo9000}, a logistic activation constrain on offset is introduced to limit the predicted bounding box near the original anchor. This makes the network more stable and increases mAP by 4.8\%.

\subsection{YOLOv3}
YOLOv3 \cite{redmon2018yolov3} is another improved detector based on YOLOv2. The major improvement is on the convolutional network. Based on the idea of Feature Pyramid Networks \cite{lin2017feature}, YOLOv3 predicts the boxes at 3 different scales, which helps to detect small objects. The idea of skip connection, as discussed in the following section, is also added into the design. As a result, the network becomes larger, with 53 layers compared to its original 19 layers. Thus, it is called Darknet-53  \cite{redmon2018yolov3}. Another achievement is that Darknet-53 runs at the highest measured floating point operation speed, which is an indication that the network is better at utilizing GPU resources. 

\subsection{Single Shot Detector (SSD)}
\begin{figure}
	\includegraphics[height=125pt]{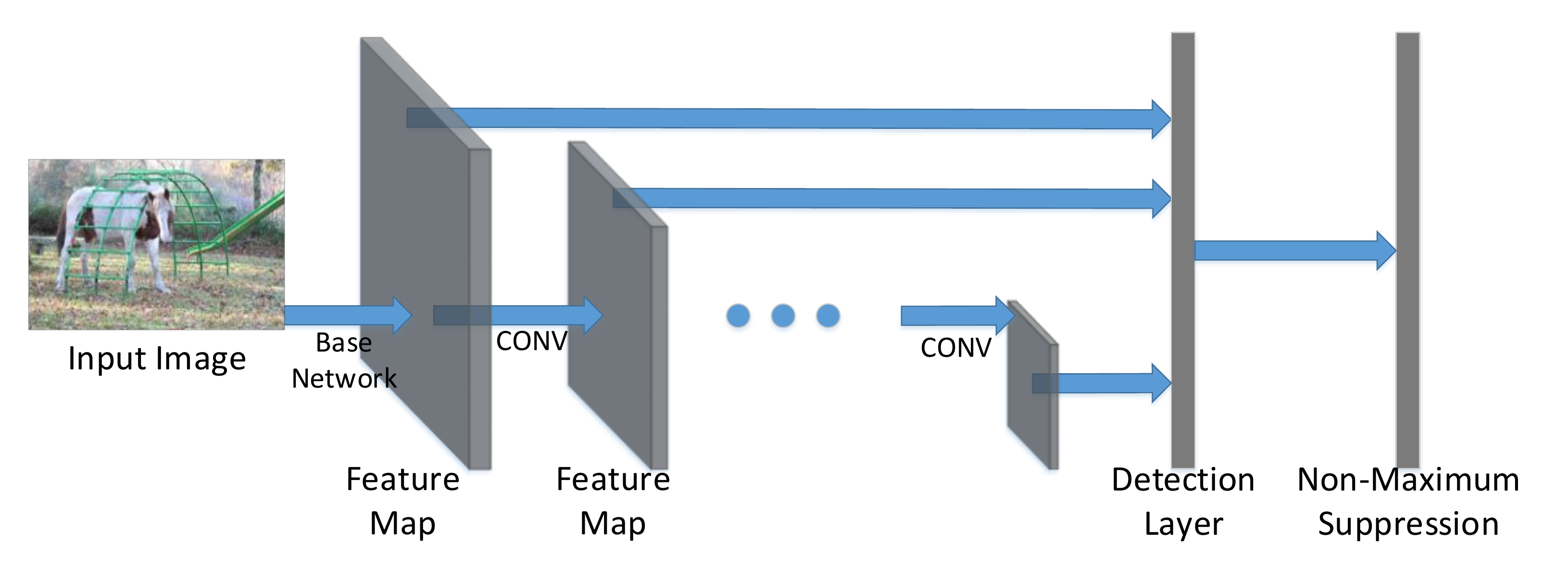}
	\centering
	\caption{\textbf{SSD architecture.} First CONV layer, which is called the base network, is truncated from a standard network. The detection layer computes confident scores for each class and offsets to default boxes.}
	\label{ssdArch}
\end{figure}

SSD \cite{liu2016ssd} was introduced to make one-stage detector run in real-time with comparative accuracy to the region proposal detectors. It is faster than YOLO \cite{redmon2016you} while generates competitive accuracy compared to the latest two-stage models like Faster R-CNN \cite{ren2015faster}. The architecture of SSD is shown in Figure~\ref{ssdArch}, which has three distinctive features. First, it uses multi-scale feature maps. From Figure~\ref{ssdArch}, we can observe that the outputs layers decrease in size progressively after a truncated base network. The layers are chosen to be output and perform detection at different scales. It allows prediction to be made at different scales. The shallower layers with more details yield better results for smaller objects, while deeper layers with information about the background are suited for larger objects. Second, the network in SSD is fully convolutional, unlike YOLO \cite{redmon2016you} which employs fully connected layers at the end. The third is the default bounding boxes and aspect ratios. In SSD, the image is divided into grid cells. Each cell in the feature maps associates a set of default bounding boxes and aspect ratios. SSD then computes category confidence scores and bounding box offsets to those default bounding boxes for each set. During prediction, SSD performs detection for objects with different sizes on the feature maps with various scales.

One-stage models inherently have the advantage of speed. In recent years, models with more than 150 fps have been published while a fast two-stage model \cite{ren2015faster} only achieves around 20 fps. To make the prediction more accurate, networks combines ideas from both one-stage and two-stage models to find the balance between speed and accuracy. For example, Faster RCNN \cite{ren2015faster} resembles one-stage model in sharing one major CNN network. One-stage models normally produce prediction from the entire feature map. So they are good at taking context information into consideration. However, this also means they are not very sensitive to smaller objects. Methods like deconvolutional single shot detector (DSSD) \cite{fu2017dssd} and RetinaNet \cite{lin2018focal} are proposed to fix this problem from different viewpoints. DSSD learned from FPN \cite{lin2017feature} modifies its CNN to predicts objects at different scales. RetinaNet develops a unique loss function to focus on hard objects. A detailed comparison results of different models on PASCAL VOC dataset are given in  Table~\ref{table:vocTable}. 
\begin{table}
	\centering
	\small
	\caption{Detection results on PASCAL VOC 2007 test set}
	\begin{tabular}{ | c | c | r | }
		\hline
		Detector & mAP (\%) & FPS \\ 
		\hline
		R-CNN \cite{girshick2014rich}                  & 58.5 & 0.02 \\  
		Fast R-CNN \cite{girshick2015fast}             & 70.0 & 0.5 \\  
		Faster R-CNN (with VGG16) \cite{ren2015faster} & 73.2 & 7 \\    
		Faster R-CNN (with ZF) \cite{ren2015faster}    & 62.1 & 18 \\  
		\hline
		YOLO \cite{redmon2016you}                & 63.4 & 45 \\  
		Fast YOLO \cite{redmon2016you}           & 52.7 & 155 \\  
		YOLO (with VGG16) \cite{redmon2016you}   & 66.4 & 21 \\    
		YOLOv2 288x288 \cite{redmon2017yolo9000} & 69.0 & 91 \\  
		YOLOv2 352x352 \cite{redmon2017yolo9000} & 73.1 & 81 \\ 
		YOLOv2 416x416 \cite{redmon2017yolo9000} & 76.8 & 67 \\ 
		YOLOv2 480x480 \cite{redmon2017yolo9000} & 77.8 & 59 \\ 
		YOLOv2 544x544 \cite{redmon2017yolo9000} & 78.6 & 40 \\
		\hline
	\end{tabular}
\label{table:vocTable}
\end{table}

\section{Other Detector Architectures}\label{detector development}
As we discussed at the beginning of this chapter, emerging detectors published after 2016 mainly focus on the performance improvement of detection. Some focus on optimizing feature extraction ability of backbone networks \cite{he2016deep, szegedy2016rethinking, huang2017densely, lin2017feature}. Some methods improve performance by modifying certain metrics, such as the loss function in RetinaNet \cite{lin2018focal} and a new IoU defination given in GIoU \cite{rezatofighi2019generalized}. With a new measuring metrics introduced in \cite{lin2014microsoft} and a comprehensive analysis performed after 2017, detection for objects with different scales becomes a major focus \cite{lin2017feature, zhou2018scale, zhang2018single, zhao2019M2DetAS}. 

In this section, we are going to discuss two examples, ResNet \cite{he2016deep} and RetinaNet \cite{lin2018focal}. ResNet \cite{he2016deep} addresses some problems caused by increasing network layers, while RetinaNet \cite{lin2018focal} tries to optimize the training efficiency.

\subsection{Deep Residual Learning (ResNet)}
\begin{figure}
	\includegraphics[height=120pt]{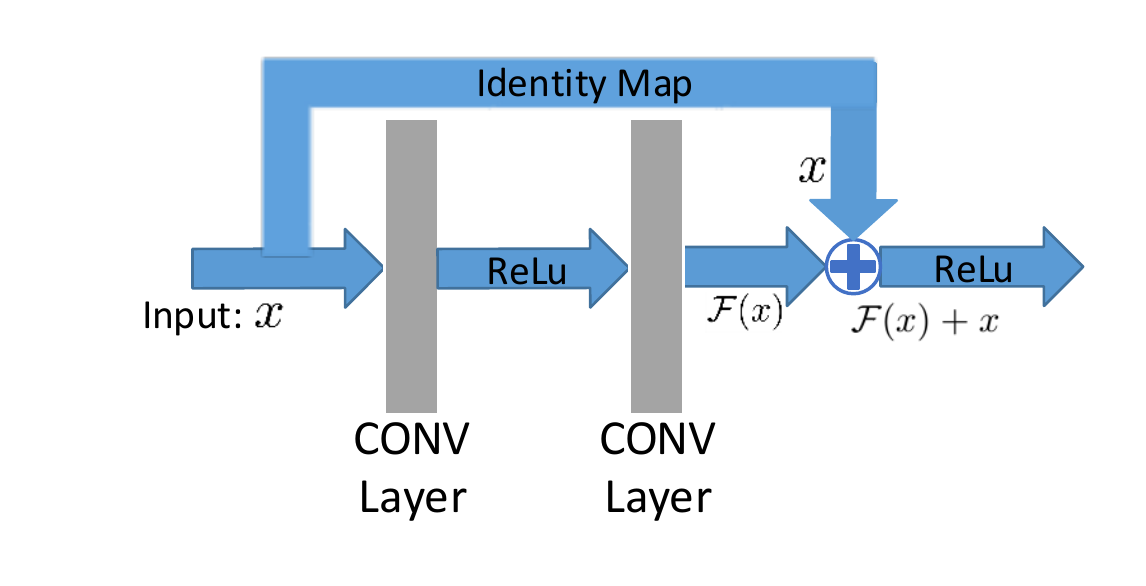}
	\centering
	\caption{\textbf{ResNet building block.} Convert the underlying mapping from \(\mathcal{H}(x)\) to \(\mathcal{F}(x)\), where \(\mathcal{H}(x)= \mathcal{F}(x)+x\).}
	\label{resBlock}
\end{figure}

Degradation is the detection accuracy drop when the network depth increases. Theoretically, this should not happen. If we assume the optimal network for a problem domain consists of certain number of layers, an even deeper network should be as good as the optimal one at least. The reason is that the added layers can be trained as identity mappings, which simply take the input and pass it to the next layer. The fact that degradation exists in practice indicates that not all systems are similarly easy to optimize \cite{he2016deep}. ResNets was published in December 2015. The effort evolves around assisting the network to find the desired mappings. The key block diagram for ResNets is shown in Figure~\ref{resBlock}.

The desired mapping without a skip connection is denoted as \(\mathcal{H}(x)= \mathcal{F}(x)+x\). By introducing the skip connection, which is an identity connection, the layers are now forced to fit the new mapping, \(\mathcal{F}(x)= \mathcal{H}(x)-x\). \(\mathcal{F}(x)\) is defined as the residual mapping. In addition, the paper hypothesizes that optimizing the residual mapping is easier than the original mapping. For example, when a layer's original optimal solution is an identity mapping, the modified layer is going to fit zero by adding the skip connection. The hypothesis is proven in the test results as discussed in next session.

Another attractive property of ResNets is that the design does not introduce any extra parameters nor computational complexity. It is desirable in two ways. First, it helps the network to achieve better performance without extra cost. Second, it facilitates fair comparison between plain networks and residual networks. During the test, we can design plain/residual networks with the same number of parameters, depth, width and computational cost \cite{he2016deep}. Therefore, the performance difference is caused just by the skip connection.

Based on a 34-layer plain network, the skip connection is added as indicated in Figure~\ref{resBlock}. When the input and output dimensions are the same, the skip connection can be added directly. And when the skip connection goes across different dimensions, the paper proposes two options: (i) Identity mapping with extra zeros paddings; and (ii) using a projection shortcut, which is the projection in the plain network block. Whichever option, a stride of 2 will be performed.

The significance of this network design is to reduce the difficulties in finding the optimal mappings of each layer. Without this design, researchers have to design architectures with different depth, train on the dataset and then compare the detection accuracy to narrow down the optimal depth. The process is very time-consuming, and different problem domains may have various optimal depth. The common training strategy, which is introduced in RCNN \cite{girshick2014rich}, is supervised pre-training on a large auxiliary dataset, followed by domain-specific fine-tuning on a small dataset. On the domain-specific dataset, it has a high chance that the network architecture is not optimal. Plus it is practically impossible to find the optimal depth for each specific problem. However, with ResNets, we can safely add more layers and expect better performance. We can rely on the networks and training to find the best performance model as long as the optimal model is shallower. Another advantage of ResNets is its efficiency, which achieves performance improvement without adding any computational complexity. 

\subsection{RetinaNet}
With all the advancements made in recent years, both two-stage methods and one-stage methods have their own pros and cons. One-stage methods are usually faster in detection, but trailing in detection accuracy. According to Lin {\it et al.} \cite{lin2018focal}, the lower accuracy is mainly caused by the extreme foreground-background class imbalance during training.  For the one-stage detector, it must sample much more candidate locations in an image. In practice, the amount of candidate locations normally goes up to around 100k covering different positions, scales and aspect ratios. Candidate locations are dominated by background examples, which makes the training very inefficient. While in two-stage detectors, most background regions are filtered out by region proposals. 

To address this issue, Lin {\it et al.} proposed RetinaNet \cite{lin2018focal}. The problem of a normal one-stage network is the overwhelming amount of easy background candidates. During training, it contributes little to the improvement of accuracy. Candidates can be easily classified as background, which means little information will be added to the model. Since the number of them is dominating, it will overpower other hard foreground candidate resulting in degenerating models. A regular cross entropy (CE) loss for binary classification is shown in Equation~\ref{retNetCE}

\begin{equation}
\label{retNetCE}
CE(p, y) = -log(p_t),{\,}where\,p_t = \left\{
  \begin{array}{ll}
    p,{\,} if\,y=1\\
    1-p,{\,}otherwise\\
  \end{array}
\right.
\end{equation}

To reduce the imbalance between the easy background and hard foreground samples, Lin {\it et al.} \cite{lin2018focal} introduced the following focal loss.

\begin{equation}
\label{retNetFL}
FL(p_t)=-(1-p_t)^\gamma log(p_t)
\end{equation}

When a sample is misclassified with a high score, \(p_t\) will have small value. Therefore, the loss is not affected that much. But when a sample is correctly classified with a high confidence, \(p_t\) is close to 1, which will significantly down-weigh its effect. In addition, a weighting factor  \(\alpha\) is employed to balance the importance of positive/negative examples.

\begin{equation}
\label{retNetFLBl}
FL(p_t)=-{\alpha}_t(1-p_t)^\gamma log(p_t)
\end{equation}

Both models in this section are addressing the problems encountered during experiments. ResNet \cite{he2016deep} has become one of the most widely used backbone networks. Its model is provided in most of the popular deep learning frameworks, such as Caffe, PyTorch, Keras, MXNet, etc. RetinaNet \cite{lin2018focal} achieves remarkable performance improvement with minimal addition in computational complexity by designing a new loss function. More importantly, it shows a new direction in optimizing the detector. Modifying evaluation metrics could result in significant accuracy increase. Rezatofighi {\it et al.} \cite{rezatofighi2019generalized} in 2019 proposed Generalized Intersection over Union (GIoU) to replace the current IoU and used GIoU as the loss function. Before the study of \cite{rezatofighi2019generalized}, all CNN detectors calculate IoU during the test to evaluate the result, while employs other metrics as loss function during training to optimize. Other recent detectors proposed after 2017 mainly focus on detecting objects at various scales, especially at small scale \cite{cai2018cascade, zhou2018scale, singh2018analysis}. Among them, the trend in design is refinement based on previous successful detectors \cite{zhang2018single, zhao2019M2DetAS}. 

\section{Performance Comparison on Different Datasets}\label{Performance}
In this section, we will present some comparative evaluation results of different models on three benchmark datasets, PASCAL VOC \cite{everingham2010pascal}, MS COCO \cite{lin2014microsoft}, and VisDrone-DET2018 \cite{zhuvisdrone2018}.

\subsection{PASCAL VOC}
\label{subsec:vocdata}
The PASCAL VOC \cite{everingham2010pascal} dataset is one of the pioneering works in generic object detection, which is designed to provide a standardized testbed for object detection, image classification, object segmentation, person layout, and action classification \cite{springer2018chapter}. The latest version is PASCAL VOC 2012. It has 20 classes and 11,530 images in the train/val dataset with 27,450 ROI annotated objects. The initial 2007 version was the first dataset with 20 classes of objects. 

To compare the performance of different networks on this dataset. We choose the PASCAL VOC 2007 testing results from models trained on PASCAL VOC 2007 + 2012. The comparison results of different models are shown in Table~\ref{table:vocTable}

\textbf{Link between one-stage and two-stage models.} From Table~\ref{table:vocTable}, we can tell the speed advantage of the one-stage detector is obvious. The first version of YOLO is almost three times faster than the fastest R-CNN detector. Looking back at the evolution of these detectors, we can see that every modifications from R-CNN leading to one-stage detectors actually fit in a trend. R-CNN has two stages, with each stage having its own separate computations. The improvements are made by eliminating duplicate calculations and combining shared features. The final version of R-CNN could be viewed as one main CNN network doing most of the work and some small regressors and CNN layers generate predictions. This architecture has become very similar to the one-stage model.

\textbf{Speed and accuracy.} For the same model, if we want to achieve higher accuracy, speed usually needs to be compromised. So it is important to find the balance for specific applications. The test results show two ways to modify. A more complex classifier or higher resolution image usually yields higher accuracy. At the same time, more time will be needed to detect the objects.

\subsection{MS COCO}
MS COCO \cite{lin2014microsoft} is a large-scale object detection, segmentation, and captioning dataset with 330k images. It aims at addressing three core problems at scene understanding: detecting non-iconic views (or non-canonical perspectives) of objects, contextual reasoning between objects and the precise 2D localization of objects \cite{lin2014microsoft}. COCO defines 12 metrics to evaluate the performance of a detector, which gives a more detailed and insight look.
Since MS COCO is a relatively new dataset, not all detectors have official results. We only compare the results of YOLOv3 and RetinaNet in Table~\ref{table:cocoTable}. Although the two models were tested under different GPUs, M40 and Titan X, their performance are almost identical \cite{redmon2018yolov3}.

\begin{table}
	\centering
	\caption{\textbf{Detection results on MS COCO.} \(AP_{50}\) and \(AP_{75}\) are average precision with IOU threshold of 0.5 and 0.75 correspondingly. \(AP\) (Average Precision) is the average over 10 thresholds ([0.5:0.05:0.95]). RetinaNet is measured on an Nvidia M40 GPU \cite{lin2018focal}, while YOLOv3 is tested on a Nvidia Titan X \cite{redmon2018yolov3}}
	\begin{tabular}{ | c | c | c | c | c | c | r | }
		\hline
		Detector & Backbone & scale(pixels) & AP & \(AP_{50}\) & \(AP_{75}\) & FPS \\ 
		\hline
		RetinaNet \cite{lin2018focal} & ResNet-101-FPN & 600 & 36.0 & 55.2 & 38.7 & 8.2 \\  
		\hline
		YOLOv3 \cite{redmon2018yolov3} & Darknet-53 & 608 & 33.0 & 57.9 & 34.4 & 19.6 \\
		\hline
	\end{tabular}
	\label{table:cocoTable}
\end{table}


\textbf{Performance difference on higher IoU threshold.} From Table~\ref{table:cocoTable}, we can see that YOLOv3 performs better than RetinaNet when IoU threshold is 50\%, while at 75\% IoU, RetinaNet has better performance, which indicates that RetinaNet has higher localization accuracy. This aligns with the effort by Lin {\it et al.} \cite{lin2018focal}. The focal loss is designed to put more weight on learning hard examples.

\subsection{VisDrone-DET2018}
\begin{figure}
	\includegraphics[height=170pt]{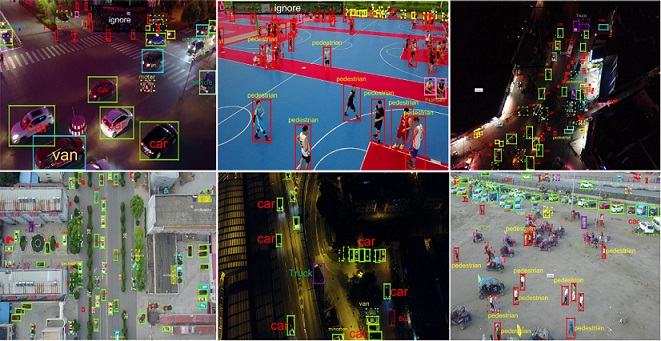}
	\centering
	\caption{Six sample images from the VisDrone-DET2018 Dataset. They are pictures taken from drones. The rectangular boxes on each image are ground truth. It has large number of small objects at distance.}
	\label{visImg}
\end{figure}

VisDrone-DET2018 dataset \cite{zhuvisdrone2018} is a special dataset that consists of images from drones. It has 8,599 images in total, including 6,471 for training, 548 for validation and 1,580 for testing. The images features a diverse real-world scenarios. The dataset was collected using various drone platforms (i.e., drones of different models), in different scenarios (across 14 different cities spanned over thousands of kilometres), and under various weather and lighting conditions \cite{springer2018chapter}. This dataset is challenging since most of the objects are small and densely populated as shown in Figure~\ref{visImg}.

To check the performance of different models on this dataset, we implement the latest version of each detector and show the result in Table~\ref{table:visTbMet} and Table~\ref{table:visTbCls}. The results are calculated on the VisDrone validation set since the test set is not publicly available.

\begin{table}
	\centering
	\small
	\caption{\textbf{Performance metrics on VisDrone-DET2018.} AR @ [maxDets=1] and [maxDets=10] mean the maximum recall given 1 detection per image and 10 detections per image correspondingly.}
	\begin{tabular}{ | c | c | l | r | }
		\hline
		Detector & iterations & AP & score \\ 
		\hline
		\multirow{7}{*}{YOLOv3 832x832} & \multirow{7}{*}{40k} & AP @ [ IoU=0.50:0.05:0.95 $|$ maxDets=500 ] & 18.74 \\
			                            &                      & AP @ [ IoU=0.50 $|$ maxDets=500 ]           & 29.77 \\
			                            &                      & AP @ [ IoU=0.75 $|$ maxDets=500 ]           & 19.53 \\
			                            &                      & AR @ [ IoU=0.50:0.05:0.95 $|$ maxDets=1 ]   & 0.91  \\
			                            &                      & AR @ [ IoU=0.50:0.05:0.95 $|$ maxDets=10 ]  & 5.63  \\
			                            &                      & AR @ [ IoU=0.50:0.05:0.95 $|$ maxDets=100 ] & 27.16 \\
			                            &                      & AR @ [ IoU=0.50:0.05:0.95 $|$ maxDets=500 ] & 27.42 \\
		\hline
		\multirow{7}{*}{Faster R-CNN}   & \multirow{7}{*}{80k} & AP @ [ IoU=0.50:0.05:0.95 $|$ maxDets=500 ] & 24.33 \\
			                            &                      & AP @ [ IoU=0.50 $|$ maxDets=500 ]           & 39.74 \\
			                            &                      & AP @ [ IoU=0.75 $|$ maxDets=500 ]           & 24.78 \\
			                            &                      & AR @ [ IoU=0.50:0.05:0.95 $|$ maxDets=1 ]   & 0.73  \\
			                            &                      & AR @ [ IoU=0.50:0.05:0.95 $|$ maxDets=10 ]  & 6.11  \\
			                            &                      & AR @ [ IoU=0.50:0.05:0.95 $|$ maxDets=100 ] & 34.26 \\
			                            &                      & AR @ [ IoU=0.50:0.05:0.95 $|$ maxDets=500 ] & 44.62 \\
			                            \hline
	\end{tabular}
	\label{table:visTbMet}
\end{table}

From Table~\ref{table:visTbMet}, it is evident that Faster R-CNN \cite{ren2015faster} performs significantly better than YOLOv3 \cite{redmon2018yolov3}. At \(IoU=0.5\), Faster R-CNN has 10\% higher accuracy. YOLO detectors inherently would struggle in datasets like this. The fully connected layers at the end take the entire feature map as input. It enables YOLO to have enough contextual information. However, it lacks local details. In addition, each grid cell can only predict a certain amount of objects, which is determined before training starts. So they have a hard time in the small and densely populated VisDrone-DET2018 dataset. It can be noted that the gap is reduced to 5\% at \(IoU=0.75\), which means YOLOv3 is catching up in terms of localization precision. We suggest this is the result of adding prediction at 3 different scales in YOLOv3 and detecting on high resolution ($832 \times 832$) images.

Table~\ref{table:visTbCls} shows the mAPs for each class from RetinaNet \cite{lin2018focal}. It shows recent detectors still need to be improved for small and morphologically similar objects. VisDrone dataset is a unique dataset, which has many potential real-life applications. Actually, most practical applications have to face different challenging situations, like bad exposure, lack of lighting, saturated number of objects. More investigation has to be done to develop more effective models to handle these complex real-life applications.
\begin{table}
	\centering
	\caption{RetinaNet on VisDrone-DET2018 without tuning.}
	\begin{tabular}{ | c | c | c | c | c | }
		\hline
		pedestrian & person & bicycle & car  & van  \\
		\hline
		12.6       & 3.6    & 4.6     & 50.2 & 21.0 \\ 
		\hline
		truck & tricycle & awning-tricycle & bus  & motor \\
		\hline
		17.8  & 10.7     & 4.9             & 32.1 & 11.7  \\
		\hline
	\end{tabular}
	\label{table:visTbCls}
\end{table}

The evolution of object detection is partially linked to the availability of labeled large-scale datasets. The existence of various datasets helps train the neural network to extract more effective features. The different characteristics of datasets motivate researchers to focus on different problems. In addition, new metrics proposed by dataset help us to better evaluate the detector performance. For example, MS COCO style mAP \cite{lin2014microsoft} helps us to understand the performance related to localization accuracy.

\section{Conclusion}\label{conclusion}
In this chapter, we have made a brief review of CNN-based object detection by presenting some of the most typical detectors and network architectures. The detector starts with a two-stage multi-step architecture and evolves to a simpler one-stage model. In latest models, even two-stage ones employ an architecture that shares a single CNN feature map so as to reduce the computational load \cite{cai2018cascade}. Some models gain accuracy increase by fusing different ideas into one detector \cite{chen2017dual, zhang2018single}. In addition to the direct effort on detectors, training strategy is also an important factor to produce high-quality results \cite{girshick2014rich, bergstra2011algorithms}. With recent development, most detectors have decent performance in both accuracy and efficiency. In practical applications, we need to make a trade-off between the accuracy and the speed by choosing a proper set of parameters and network structures. Although great progress has been made in object detection in the past years, some challenges still need to be addressed, like occlusion \cite{wu2005detection} and truncation. In addition, more well designed datasets, like VisDrone, need to be developed for specific practical applications.


\printindex

\end{document}